\documentclass{article}

\usepackage{microtype}
\usepackage{graphicx}
\usepackage{subcaption}
\usepackage{booktabs} 
\usepackage{xcolor}

\usepackage{hyperref}

\usepackage[preprint]{icml2026}

\usepackage{amsmath}
\usepackage{amssymb}
\usepackage{mathtools}
\usepackage{amsthm}

\usepackage{booktabs}
\usepackage{tabularx}
\usepackage{multirow}
\usepackage{xcolor}
\usepackage{pifont}
\usepackage{amssymb}
\usepackage{tikz}
\usetikzlibrary{shapes.geometric, arrows.meta, positioning, fit, backgrounds}
\newcommand{\cmark}{\textcolor{blue}{\ding{51}}}
\newcommand{\xmark}{\textcolor{red}{\ding{55}}}

\usepackage[capitalize,noabbrev]{cleveref}

\theoremstyle{plain}

\theoremstyle{definition}

\theoremstyle{remark}

\usepackage[textsize=tiny]{todonotes}
\usepackage{tcolorbox}
\tcbuselibrary{breakable}

\icmltitlerunning{ProjDevBench: Benchmarking AI Coding Agents on End-to-End Project Development}

\begin{document}

\twocolumn[
  \icmltitle{ProjDevBench: Benchmarking AI Coding Agents on \\End-to-End Project Development}



  \icmlsetsymbol{equal}{*}

  \begin{icmlauthorlist}
    \icmlauthor{Pengrui Lu}{equal,ucm,sjtu,sii}
    \icmlauthor{Shiqi Zhang}{equal,sjtu}
    \icmlauthor{Yunzhong Hou}{equal,bit}
    \icmlauthor{Lyumanshan Ye}{sjtu}
    \icmlauthor{Chaoyi Huang}{sjtu}
    \icmlauthor{Zixi Chen}{sjtu}
    \icmlauthor{Ji Zeng}{sjtu}
    \icmlauthor{Hantao Jiang}{sjtu}
    \icmlauthor{Pengfei Liu}{sjtu,sii}
    \icmlauthor{Yiwei Wang}{ucm}
    \icmlauthor{Ming-Hsuan Yang}{ucm}
  \end{icmlauthorlist}
  \icmlaffiliation{sjtu}{Shanghai Jiao Tong University, Shanghai, China}

  \icmlaffiliation{bit}{Beijing Institute of Technology, Beijing, China}

  \icmlaffiliation{ucm}{University of California, Merced, CA, USA}
  \icmlaffiliation{sii}{Shanghai Innovation Institute, Shanghai, China}

  \icmlcorrespondingauthor{Ming-Hsuan Yang}{mhyang@ucmerced.edu}
\icmlcorrespondingauthor{Yiwei Wang}{yiweiwang2@ucmerced.edu}
\icmlcorrespondingauthor{Pengfei Liu}{pengfei@sjtu.edu.cn}

  \icmlkeywords{Machine Learning, ICML}

  \vskip 0.3in
]



\printAffiliationsAndNotice{\icmlEqualContribution}

\begin{abstract}

Recent coding agents can generate complete codebases from simple prompts, yet existing evaluations focus on issue-level bug fixing and lag behind end-to-end development. We introduce \textsc{ProjDevBench}, an end-to-end benchmark that provides project requirements to coding agents and evaluates the resulting repositories.
Combining Online Judge (OJ) testing with LLM-assisted code review, the benchmark evaluates agents on (1) system architecture design, (2) functional correctness, and (3) iterative solution refinement.
We curate 20 programming problems across 8 categories, covering both concept-oriented tasks and real-world application scenarios, and evaluate six coding agents built on different LLM backends.
Our evaluation reports an overall acceptance rate of 27.38\%: agents handle basic functionality and data structures but struggle with complex system design, time complexity optimization, and resource management. Our benchmark is available at this
\href{https://github.com/zsworld6/projdevbench}{\textcolor{blue}{https://github.com/zsworld6/projdevbench}}. 



\end{abstract}
\section{Introduction}

Recent progress in large language models has enabled coding agents to participate in software development workflows that extend beyond generating individual functions or files. This allows both developers and users with no prior coding knowledge alike to provide high-level project requirements and rely on a coding agent to implement the majority of the system—a workflow sometimes referred to as \textit{vibe coding}.

\begin{figure}[t]
    \centering
    \includegraphics[width=\linewidth]{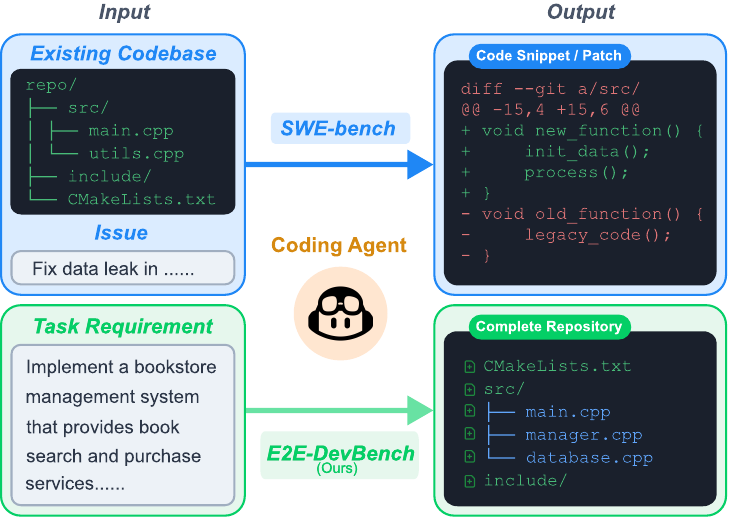} 
\caption{Task comparison. Unlike benchmarks where coding agents modify code snippets from pre-existing codebases based on issues or pull requests, our benchmark evaluates end-to-end repository construction directly from project-level requirements.}
    \label{fig:tasks}
    \vspace{-3mm}
\end{figure}

These workflows point to a growing expectation that coding agents should be capable of constructing and even executing complete, runnable software at the project level. Given only an initial natural language specification, a coding agent is expected to autonomously determine the project structure, create and organize multiple source files, configure dependencies, and ultimately deliver a fully functional system (\cref{fig:tasks} bottom). This setting emphasizes end-to-end project construction, where success depends not only on local code correctness but also on maintaining consistency across files, making coherent system-level design decisions, and ensuring the final system is fully executable.

\begin{table*}[t]
\caption{
Comparison with existing coding benchmarks.
\textbf{From Scratch}: whether including tasks that require building a project without initial code or templates.
\textbf{Output}: the granularity of expected submission.
\textbf{E2E Exec.}: whether the benchmark requires building and running the complete system.
\textbf{Diag. Feedback}: whether execution provides fine-grained diagnostic signals (e.g., error types, partial credit) beyond binary pass/fail.
\textbf{Code Review}: specification-level compliance beyond testing.
}
\centering
\small
\begin{tabular}{lccccc}
\toprule
\textbf{Benchmark} 
& \textbf{From Scratch} 
& \textbf{Output} 
& \textbf{E2E Exec.} 
& \textbf{Diag. Feedback}
& \textbf{Code Review} \\
\midrule
HumanEval / MBPP
& \xmark & Function & \xmark & \xmark & \xmark \\

APPS / CodeContests
& \xmark & Single-file & \xmark & \xmark & \xmark \\

RepoBench
& \xmark & Repo (Partial) & \xmark & \xmark & \xmark \\

SWE-bench
& \xmark & Patch & \xmark & \xmark & \xmark \\

DevEval
& \cmark & Repository & \xmark & \xmark & \xmark \\

E2EDevBench
& \cmark & Repository & \cmark & \xmark & \cmark \\

InnovatorBench
& \xmark & Research Artifacts & \cmark & \xmark & \xmark \\

NL2Repo-Bench 
& \cmark & Repository & \cmark & \xmark & \xmark \\

\midrule
\textbf{ProjDevBench (Ours)}
& \cmark & Repository & \cmark & \cmark & \cmark \\
\bottomrule
\end{tabular}
\label{tab:benchmark_comparison}
\vspace{-3mm}
\end{table*}

Despite this shift towards end-to-end project construction, \textbf{most existing benchmarks for coding agents still focus on a smaller scale}:
HumanEval \cite{chen2021codex} and MBPP \cite{austin2021program} evaluate function-level code generation, while 
SWE-bench \cite{jimenez2024swebench} and others study issue resolving in existing codebases (\cref{fig:tasks} top).
These benchmarks test an agent's ability to understand or modify existing code to a certain extent, 
yet, they cannot reflect the competency of coding agents on end-to-end development tasks, which aim to construct complete projects from high-level specifications.

To address this gap, we propose \textbf{ProjDevBench}, a benchmark designed to evaluate coding agents on \textbf{end-to-end project construction}. 
In terms of requirements or input to the coding agents, ProjDevBench includes tasks where agents receive only simple high-level instructions without any initial codebase, in contrast to existing benchmarks that provide complete repository structures \cite{jimenez2024swebench,liu2023repobench} or staged reference materials \cite{DevEval} alongside requirements. 
In terms of expected outputs from the coding agents, instead of single-file updates \cite{chen2021codex,austin2021program} or atomic fixes \cite{jimenez2024swebench}, agents are required to generate full software repositories that can be executed in practice. 
We show a detailed comparison between ProjDevBench and existing benchmarks in \cref{tab:benchmark_comparison}.
%

We employ two complementary metrics to evaluate complete repositories generated by agents. 
First, we assess functional correctness through automated execution on an Online Judge (OJ) platform, which compiles, runs, and tests submissions against comprehensive test suites, and finally provides detailed diagnostics on failure modes. 
Second, since OJ evaluation alone cannot detect rule violations or cheating solutions, we introduce a code review mechanism that combines rule-based Python scripts for explicit constraint violations (e.g., forbidden library usage) with LLM-based review for subtler compliance issues.


A diverse set of tasks drawn from realistic programming scenarios is included in our ProjDevBench, enabling systematic evaluation of coding agents on their ability to plan, implement, and integrate components at the project level, effectively pushing the frontier of autonomous software development. 
Specifically, we curate 20 problems across 8 categories from coding concepts to real-world applications. 
These tasks demand extended interaction, with agents averaging 138 interaction turns and 4.81M tokens per problem in our experiments.

%


Our evaluation on six coding agents built on different LLM backends reveals several key findings: (1) Model performance varies significantly across tasks and models: Codex \cite{openai2024codex} with GPT-5 \cite{singh2025openai} achieves the best overall performance (77.85\%), with performance gaps widening on from-scratch construction tasks; GPT-5 generally excels at execution score while Sonnet-4.5 \cite{anthropic2025claudesonnet45} shows stronger code review and specification compliance; (2) Agents struggle with multiple critical aspects: specification alignment, edge case handling, time complexity optimization, and resource management, with 42\% of failures attributed to wrong answers and 14\% to time limits; 
(3) Extended interaction indicates difficulty and correlates negatively with performance, suggesting agents struggle to convert prolonged debugging into progress.

The contributions of our work are as follows:

\begin{itemize}
    \item We introduce ProjDevBench, a benchmark for evaluating coding agents on end-to-end project construction, which includes autonomous design, build configuration, and strict performance constraints.
    \item We establish a dual evaluation protocol combining Online Judge (OJ) for execution-based correctness assessment with LLM-assisted code review for rule violations and cheating solutions detection.
    \item We provide comprehensive empirical analysis on 6 coding agents across 6 LLM backends, revealing systematic failure modes in specification alignment, edge case handling, complexity optimization, resource management, and multi-turn interaction patterns.
\end{itemize}




\begin{figure*}[ht]
\centering
\vspace{-1mm}
\includegraphics[width=1\linewidth]{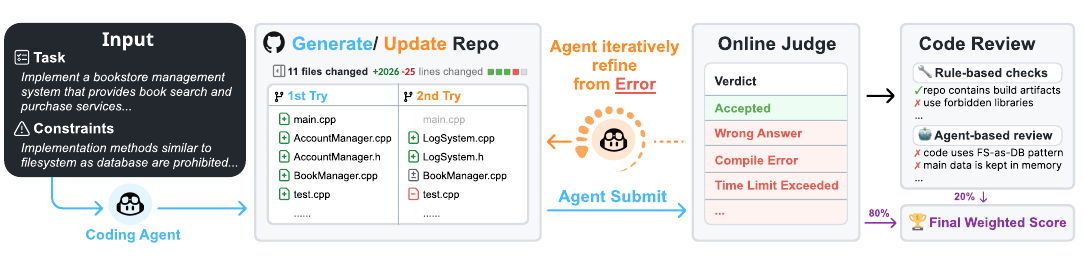}
\vspace{-7mm}
\caption{Overview of the benchmark pipeline.}
\label{fig:benchmark_pipeline}
\vspace{-3mm}
\end{figure*}


\section{Related Work}
\label{sec:related}

\paragraph{AI Coding Agents.} 
Modern coding agents span a spectrum of autonomy and integration modes.
\textbf{IDE-integrated assistants} such as GitHub Copilot \cite{github2024copilot} and Cursor \cite{cursor2024} provide real-time code suggestions within development environments, with newer versions incorporating increasing agent capabilities for multi-file editing and contextual understanding.
\textbf{Autonomous frameworks} such as OpenHands \cite{wang2025openhandsopenplatformai} and CodeAgent \cite{zhang-etal-2024-codeagent} can independently operate terminal tools, manage file systems, and execute multi-step workflows.
Recent research has explored specialized agent-computer interfaces for software engineering \cite{yang2024sweagent}, simpler non-agentic approaches \cite{agentless}, and autonomous program improvement \cite{zhang2024autocoderoverautonomousprogramimprovement}.
While these agents increasingly enable vibe coding, their ability to handle complex system-level decisions, build configurations, and from-scratch project construction remains under-explored in structured evaluations.

\paragraph{Evolution of AI-Assisted SE Tasks.} 
Software engineering tasks for LLMs have shifted in granularity. Initial research focused on \textbf{function-level synthesis} and \textbf{repository-level completion} \cite{zhang-etal-2023-repocoder,liu2023repobench}, where models fill in missing logic within a predefined context. More recently, the focus has moved to \textbf{software maintenance}, such as bug fixing in SWE-bench \cite{jimenez2024swebench}, which requires generating patches for existing code. Agent frameworks have been enhanced through reasoning-acting paradigms \cite{yao2023react}, executable code actions \cite{wang2024executable}, and iterative self-refinement \cite{10.5555/3666122.3668141}. However, \textbf{end-to-end project construction} remains a distinct challenge. Unlike maintenance, it requires agents to autonomously design directory structures, manage inter-file dependencies, configure build systems (e.g., CMake) without a pre-existing template, and autonomously run the project for testing. Crucially, such tasks demand \textbf{extended agent-environment interaction}---iterative cycles of code generation, execution, and refinement---that single-shot or limited-turn evaluations cannot capture.

\paragraph{Code Generation Benchmarks.} 

Existing benchmarks primarily target specific stages of the development cycle but fall short of evaluating holistic project construction. Early benchmarks like HumanEval \cite{chen2021codex} and MBPP \cite{austin2021program} are limited to single-function snippets. Competition-level benchmarks such as APPS \cite{hendrycksapps2021} and CodeContests \cite{li2022competition} evaluate algorithmic problem-solving but remain single-file tasks. Repository-level benchmarks are often patch-based, such as SWE-bench \cite{jimenez2024swebench}, which evaluates bug fixing in existing codebases, and RepoBench \cite{liu2023repobench}, which focuses on code completion within existing repositories. 
Moreover, most existing benchmarks adopt single-shot evaluation, where a model generates code in one pass without iterative refinement.

Recent work has begun exploring end-to-end evaluation. DevEval \cite{DevEval}
decomposes development into staged tasks with reference inputs such as UML
diagrams provided at each phase, but does not require fully autonomous
construction. E2EDevBench \cite{zeng2025benchmarkingstudyingllmbasedagent}
evaluates agents on software development with hybrid test-case and LLM-based
verification, focusing on PyPI package development. NL2Repo-Bench
\cite{ding2025nl2repo} evaluates long-horizon repository generation from
natural-language requirements, where agents must produce fully installable
Python libraries validated against upstream pytest suites. InnovatorBench
\cite{wu2025innovatorbenchevaluatingagentsability} assesses agents on ML
research automation with provided templates and scaffolds. However, these
benchmarks primarily provide binary pass/fail feedback and focus on narrow domains, limiting their ability to capture diverse
real-world software engineering challenges, including build system
configuration, strict resource constraints, and complex system-level design.

\section{ProjDevBench}

As shown in \cref{fig:benchmark_pipeline}, we design an end-to-end benchmark pipeline for coding agents, where an agent is given task descriptions and constraints in natural language, and is required to autonomously construct and iteratively refine a complete code repository. Each submission is evaluated through execution on an Online Judge, together with rule-based and LLM-based code review, and all evaluation signals are aggregated into a final weighted score.

\subsection{Task Definition and Scope}



\begin{figure*}[ht]
\centering
\includegraphics[width=0.8\linewidth]{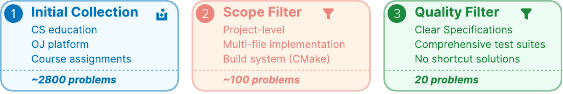}
\caption{
Problem collection and filtering. 
}
\label{fig:problem_collection}
\vspace{-2mm}
\end{figure*}

\paragraph{Model input.}

For each task, the agent is provided with a problem specification written in natural language.
The specification describes the expected functionality, input and output formats, constraints, and submission requirements. 
For tasks in the \textit{project-completion setting}, a partial codebase is provided as part of the input, representing an incomplete project that must be completed; for tasks in the \textit{project-creation setting}, no initial codebase is provided, and the agent is required to implement the entire project from scratch based solely on the specification.

While our primary focus is on evaluating an agent’s ability to construct complete software projects from scratch, we include both settings to enable controlled comparisons between project creation and project completion under a unified project-level evaluation objective; for analysis purposes, tasks in the project-completion and project-creation settings are categorized as \textit{Easy} and \textit{Hard} subsets, respectively.


\paragraph{Model output.}
The primary output of an agent is a software project rather than a code fragment.
For most tasks, this output takes the form of a Git repository containing multiple source files, a valid build configuration (e.g., a \texttt{CMakeLists.txt}), and an executable that can be compiled and run successfully.
Thus, correctness depends not only on implementing the required logic, but also on producing a coherent and buildable project that satisfies all task requirements.
In fact, human reference solutions are non-trivial, typically consisting of multiple interacting files, with an average of around ten source files per project.



\subsection{Problem Collection}

We curate problems for ProjDevBench using a three-stage pipeline designed to emphasize end-to-end software construction rather than isolated algorithmic tasks.

\paragraph{Stage I: Initial collection.}
We collect approximately 2,800 candidate problems from a large-scale Online Judge platform used in undergraduate computer science education.
This platform hosts both traditional algorithmic problems similar to competitive programming and course assignments that require building complete, multi-file software systems.

\paragraph{Stage II: Scope-based filtering.}
We retain problems that involve project-level development, such as multi-file implementations, explicit module organization, build system configuration (e.g., \texttt{CMakeLists.txt}), reusable abstractions, or stateful multi-command interfaces.
Purely algorithmic or single-function tasks are excluded, reducing the candidate set to approximately 100 problems.

\paragraph{Stage III: Quality-based filtering.}
We further select problems with clear specifications, comprehensive test suites covering both functionality and edge cases, and non-trivial difficulty as reflected by historical submission outcomes.
Problems with test vulnerabilities or shortcut solutions are removed, yielding the final set of 20 problems.

\begin{figure}[t]
\centering
\includegraphics[width=1\linewidth]{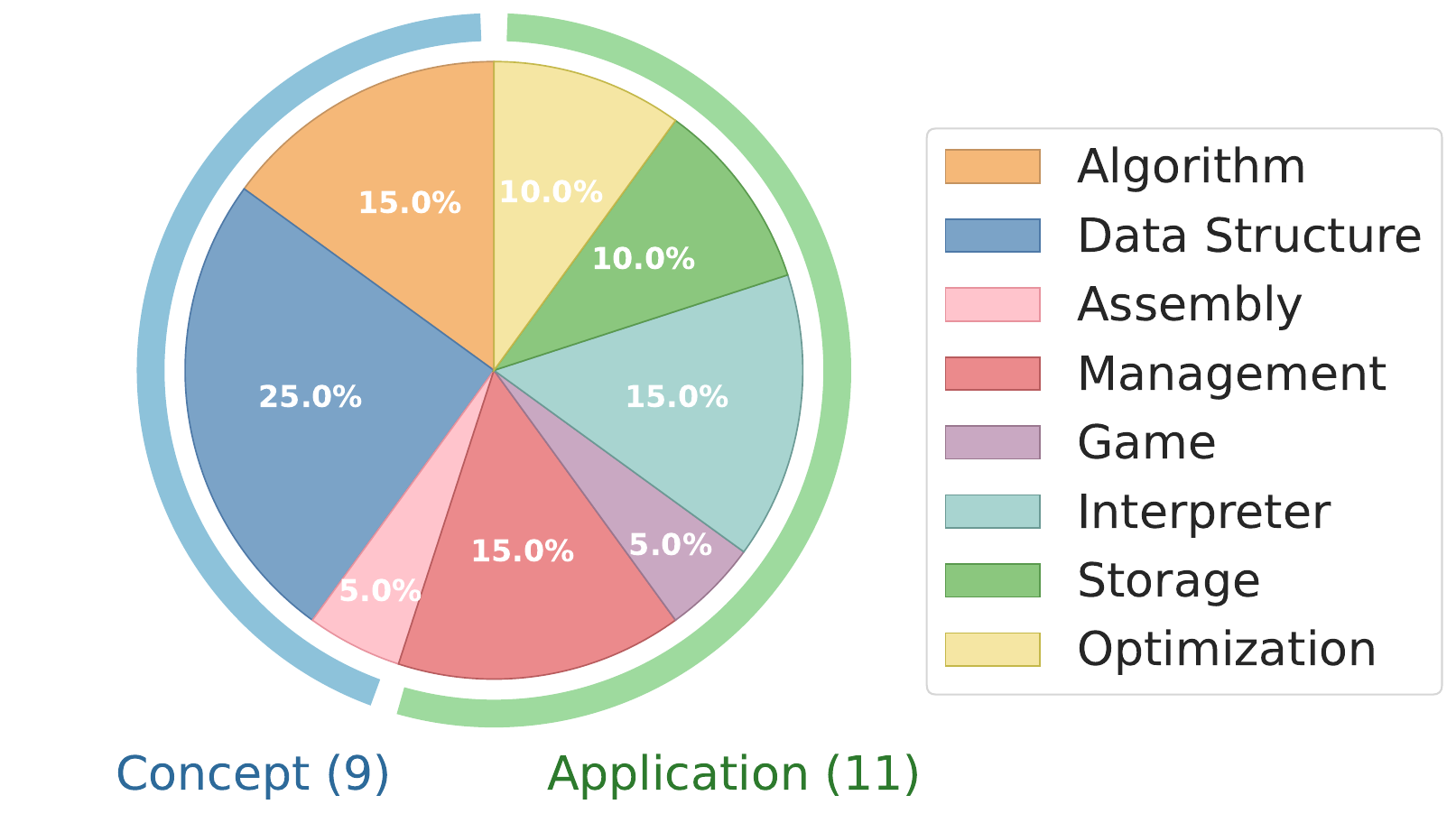}
\vspace{-4mm}
\caption{Distribution of ProjDevBench tasks across 8 categories.}
\label{fig:category_pie_chart}
\vspace{-3mm}
\end{figure}

Through this process, we obtain 20 problems covering a range of realistic software development scenarios, including data structures, interpreters, management systems, and storage components.
Domain experts then reformulate the original problem statements into standardized task descriptions suitable for agent evaluation.
Detailed task-level statistics and problem attributes are provided in \cref{tab:problem_details}, with the distribution of task categories summarized in \cref{fig:category_pie_chart}.

\subsection{Evaluation Protocol}

ProjDevBench adopts a dual evaluation protocol that distinguishes between hard functional correctness and compliance at the rule level and specification level.


\paragraph{Execution-based evaluation.}
All submissions are evaluated using comprehensive test suites on the Online Judge platform.
Test cases verify end-to-end executability, functional correctness, edge-case handling, and compliance with problem-specific time and memory limits.
Each submission receives an \emph{Execution Score} computed as a weighted sum of passed test cases, where individual test points may carry different score weights reflecting their relative importance or difficulty.
In contrast to binary pass/fail evaluation in prior work \cite{zeng2025benchmarkingstudyingllmbasedagent}, the Online Judge provides \emph{fine-grained verdict-level failure signals}—including compile errors, runtime errors, time limit exceeded, memory limit exceeded, and wrong answer verdicts—that enable systematic diagnosis of agent failure modes and support iterative debugging, as demonstrated in \cref{sec:analysis}.

\paragraph{Code review.}
In addition to OJ testing, ProjDevBench performs code review to assess whether a submission genuinely follows the problem specification.
This evaluation focuses on criteria that cannot be reliably captured by test cases, including violations of explicit problem rules (e.g., forbidden libraries), hack-based or adversarial solutions, and implementations that exploit weaknesses in the test suite rather than adhering to the intended constraints.
Each submission receives a \emph{Code Review score} reflecting its compliance with these requirements.
We employ an LLM-based code review approach to judge such specification-level compliance.

\paragraph{Final scoring.}
The final score is computed as a weighted combination of the OJ score and the Code Review score, with functional correctness accounting for the majority of the weight.
This design prioritizes producing correct and executable solutions while penalizing submissions that violate essential requirements or problem constraints.

\section{Experiments}

\begin{table*}[ht]
\caption{
Performance on ProjDevBench (``Exec.'': Execution; ``CR'': Code Review; ``Final'': Weighted Score)
}
\centering
\small
\begin{tabular}{llccccccccc}
\toprule
\multirow{2}{*}{\textbf{Agent}} & \multirow{2}{*}{\textbf{Model}} & \multicolumn{3}{c}{\textbf{Easy (E)}} & \multicolumn{3}{c}{\textbf{Hard (H)}} & \multicolumn{3}{c}{\textbf{Overall}} \\
\cmidrule(lr){3-5} \cmidrule(lr){6-8} \cmidrule(lr){9-11}
& & \textbf{Exec.} & \textbf{CR} & \textbf{Final} & \textbf{Exec.} & \textbf{CR} & \textbf{Final} & \textbf{Exec.} & \textbf{CR} & \textbf{Final} \\
\midrule
\multirow{2}{*}{Augment} & GPT-5 & 77.10 & 76.00 & 76.88 & 57.22 & 65.03 & 58.78 & 72.13 & 73.26 & 72.35 \\
& Sonnet-4.5 & 69.14 & \textcolor{blue}{\textbf{92.56}} & 73.83 & 56.81 & 67.43 & 58.93 & 66.06 & 86.28 & 70.10 \\
\midrule
\multirow{2}{*}{Codex} & GPT-5 & \textcolor{blue}{\textbf{79.24}} & 82.11 & \textcolor{blue}{\textbf{79.81}} & \textcolor{blue}{\textbf{69.22}} & 82.90 & \textcolor{blue}{\textbf{71.95}} & \textcolor{blue}{\textbf{76.73}} & 82.31 & \textcolor{blue}{\textbf{77.85}} \\
& Sonnet-4.5 & 66.07 & 68.22 & 66.50 & 31.88 & 83.23 & 42.15 & 57.52 & 71.98 & 60.41 \\
\midrule
Gemini CLI & Gemini-3-Pro-Preview & 74.57 & 80.33 & 75.72 & 35.53 & \textcolor{blue}{\textbf{94.20}} & 47.26 & 64.81 & 83.80 & 68.61 \\
\midrule
\multirow{3}{*}{Cursor} 
& GPT-5 & 69.74 & 80.56 & 71.90 & 67.80 & 87.27 & 71.69 & 69.26 & 82.23 & 71.85 \\
& Sonnet-4.5 & 71.12 & 85.67 & 74.03 & 60.17 & 66.47 & 61.43 & 68.39 & 80.87 & 70.88 \\
& Gemini-3-Pro-Preview & 72.87 & 88.67 & 76.03 & 71.47 & 80.03 & 73.18 & 72.52 & 86.51 & 75.32 \\
\midrule
\multirow{3}{*}{GitHub Copilot} 
& GPT-5 & 59.79 & 71.11 & 62.06 & 55.20 & 82.90 & 60.74 & 58.64 & 74.06 & 61.73 \\
& Sonnet-4.5 & 71.10 & 87.89 & 74.46 & 36.63 & 80.23 & 45.35 & 62.48 & 85.97 & 67.18 \\
& Gemini-3-Pro-Preview & 59.22 & 58.11 & 59.00 & 36.44 & 75.67 & 44.29 & 53.53 & 62.50 & 55.32 \\
\midrule
\multirow{5}{*}{Claude Code} & DeepSeek-V3.2-Exp & 50.05 & 60.78 & 52.20 & 35.23 & 82.77 & 44.74 & 46.34 & 66.28 & 50.33 \\
& GLM-4.6 & 56.25 & 80.89 & 61.18 & 39.22 & 84.37 & 48.25 & 52.00 & 81.76 & 57.95 \\
& GPT-5 & 50.69 & 84.33 & 57.42 & 50.59 & 83.13 & 57.10 & 50.67 & 84.03 & 57.34 \\
& Kimi-k2-0905-Preview & 53.49 & 65.89 & 55.97 & 35.56 & 73.67 & 43.18 & 49.00 & 67.83 & 52.77 \\
& Sonnet-4.5 & 66.85 & 92.89 & 72.06 & 54.47 & 78.57 & 59.29 & 63.76 & \textcolor{blue}{\textbf{89.31}} & 68.87 \\
\bottomrule
\end{tabular}
\label{tab:agent_difficulty_results}
\vspace{-2mm}
\end{table*}

\subsection{Experimental Setup}

\paragraph{Agents and Models.}
We evaluate six coding agents via their command-line interfaces: \textbf{Cursor} \cite{cursor2024}, \textbf{GitHub Copilot} \cite{github2024copilot}, \textbf{Claude Code} \cite{anthropic2024claudecode}, \textbf{Augment} \cite{augment2024}, \textbf{Codex CLI} \cite{openai2024codex}, and \textbf{Gemini CLI} \cite{gemini-cli}.
We test agents with frontier models \textbf{GPT-5} \cite{singh2025openai}, \textbf{Claude Sonnet 4.5} \cite{anthropic2025claudesonnet45}, and \textbf{Gemini 3 Pro Preview} \cite{google2025gemini3}. 
GPT-5 and Sonnet-4.5 are evaluated across five agents, while Gemini 3 Pro is evaluated on Cursor, Copilot, and Gemini CLI. 
For Claude Code, we additionally evaluate open-source models \textbf{GLM-4.6} \cite{glm2024chatglm}, \textbf{Kimi-k2-0905-Preview} \cite{team2025kimi}, and \textbf{DeepSeek-V3.2-Exp} \cite{liu2025deepseek}.

\paragraph{Implementation details.}
For each agent-model configuration, we run a single evaluation pass on every problem in the benchmark.
Each run is constrained by the maximum number of submissions allowed per problem (ranging from 2 to 18 depending on problem complexity), rather than a fixed time budget.
All agents are evaluated using the same prompts and problem specifications to ensure fair comparison.
The final score weighs execution correctness at 80\% and code review compliance at 20\%.




\subsection{Main Results}

\cref{tab:agent_difficulty_results} presents the execution, code review, and final weighted scores for each configuration. Our results answer the following three key questions.

\paragraph{Which agent framework achieves the best overall performance?}
Codex paired with GPT-5 achieves the highest final weighted score of 77.85, outperforming Augment+GPT-5 at 72.35, Cursor+GPT-5 at 71.85, and Claude Code+Sonnet-4.5 at 68.87. This advantage stems primarily from its coding capabilities, where Codex+GPT-5 reaches 76.73 execution score compared to Augment+GPT-5's 72.13. The difference becomes more pronounced on hard problems requiring from-scratch construction, where Codex+GPT-5 maintains 69.22 execution score versus Augment+GPT-5's 57.22 and Cursor+GPT-5's 67.80. Notably, Cursor+Gemini-3-Pro-Preview achieves 75.32, demonstrating competitive performance. Both Cursor and Augment demonstrate stable performance across different base models, with all their tested configurations achieving final scores above 70, suggesting their framework design is relatively stable regardless of the underlying language model choice.

\paragraph{How do different base models affect agent performance?}
Model selection exhibits notable interactions with agent framework design. For execution scores, GPT-5 generally outperforms Sonnet-4.5 across most frameworks, with the gap varying by framework. In Codex, GPT-5 achieves 76.73 versus Sonnet-4.5's 57.52, while in Augment, GPT-5 reaches 72.13 compared to Sonnet-4.5's 66.06. Gemini-3-Pro-Preview shows mixed results in execution score: it achieves 72.52 in Cursor (comparable to GPT-5's 69.26) but only 53.53 in GitHub Copilot. The pattern differs for code review compliance: In Claude Code, Sonnet-4.5 achieves the highest code review score of 89.31, exceeding GPT-5's 84.03, while Augment+Sonnet-4.5 scores 86.28 compared to Augment+GPT-5's 73.26. Within the Claude Code framework, we also evaluate three open-source models, which achieve final scores of 52.77, 50.33, and 57.95 respectively. GLM-4.6 achieves the highest final score at 57.95, slightly surpassing GPT-5 at 57.34, while Kimi-k2-0905-Preview and DeepSeek-V3.2-Exp lag notably behind at 52.77 and 50.33. All open-source models remain behind Sonnet-4.5's 68.87 final score, indicating a performance gap between open-source and the strongest closed-source models.

\paragraph{How does problem difficulty affect agent robustness?}
The transition from easy problems with partial codebases to hard problems requiring from-scratch construction reveals differences in framework robustness. Codex+GPT-5 maintains relatively stable execution performance, declining from 79.24 on easy problems to 69.22 on hard. Cursor+GPT-5 demonstrates similar stability with a drop from 69.74 to 67.80. In contrast, GitHub Copilot+Sonnet-4.5 shows substantial degradation, decreasing from 71.10 to 36.63, while Gemini CLI drops sharply from 74.57 to 35.53. Code review scores exhibit varied patterns: Cursor+GPT-5 increases from 80.56 to 87.27, and Codex+Sonnet-4.5 rises from 68.22 to 83.23, suggesting that specification compliance issues may manifest differently when agents modify existing codebases compared to constructing projects from scratch, potentially due to the complexity of understanding and integrating with pre-existing code structures and architectural decisions.

\section{Analysis and Discussion}
\label{sec:analysis}
\subsection{Where End-to-End Coding Agents Fail}

\cref{tab:submission_status} summarizes submission outcomes across all agents. Only 27.38\% of submissions were accepted, with the majority 
failing due to wrong answers (41.86\%) or time limit violations (13.91\%). 
Rather than analyzing these statuses in isolation, we group them into higher-level 
failure modes based on their underlying causes, as discussed below.

\begin{table}[t]
\caption{Distribution of submission status types across all agents.}
\centering \small
\begin{tabular}{lcc}
\toprule
\textbf{Status Type} & \textbf{Count} & \textbf{Percentage} \\
\midrule
Accepted & 484 & 27.38\% \\ \midrule
Wrong Answer & 740 & 41.86\% \\
Time Limit Exceeded & 246 & 13.91\% \\
Runtime Error & 124 & 7.01\% \\
Compile Error & 80 & 4.52\% \\
Memory Leak & 62 & 3.51\% \\
Memory Limit Exceeded & 24 & 1.36\% \\
Others & 8 & 0.45\% \\
\bottomrule
\end{tabular}
\label{tab:submission_status}
\vspace{-2mm}
\end{table}

\paragraph{Specification misalignment.} Agents frequently fail due to misalignment between their understanding of problem specifications and actual requirements, evidenced by Wrong Answer and Compile Error submissions. Functional incompleteness appears when agents generate syntactically correct frameworks but omit critical business logic—in Train Ticket Management, all agents omitted the seat management system despite implementing user management and train querying. Similarly, in Minesweeper, agents accessed only 3,789 of 3,825 safe blocks, indicating incomplete implementation rather than logical errors. Structural misunderstanding manifests when agents fail to distinguish development from submission contexts: in int2048, agents included test code with \texttt{main()} functions, causing redefinition errors. Algorithmic understanding deviations also occur, as in Mini-Aidiv-N where agents implemented Softmax by summing entire matrices rather than row-wise, resulting in 0.02 accuracy despite runnable code.

\paragraph{Edge case handling deficiencies.} Agents demonstrate systematic weaknesses in boundary condition handling, leading to both Wrong Answer and Runtime Error failures. Bookstore Hidden Test Points saw all agents fail, indicating struggles with edge cases like empty strings, file I/O exceptions, and nested scenarios. Runtime safety failures manifest at the test case level, with numerous Segmentation Faults and Aborted errors caused by null pointer dereferences and array bounds violations. In map implementation, Red-Black Tree implementations lacked proper null checks in rotation functions, while ICPC Management System had many test cases resulting in Aborted errors from uninitialized variables. Detail processing errors also appear: in Bookstore, agents used substring matching instead of exact keyword matching, causing ``math'' to incorrectly match ``mathematics''.


\paragraph{Time complexity analysis deficiencies.}
Agents exhibit systematic weaknesses in time complexity reasoning, resulting in a substantial number of Time Limit Exceeded submissions despite functionally correct implementations. In the ICPC Management System, agents re-sorted all teams after each unfreeze operation, yielding an $O(K \times N \log N)$ solution, whereas the correct approach exploits the locality of ranking changes to achieve $O(K \log N)$ using ordered data structures. This failure reflects an inability to identify and leverage problem-specific structural properties.

More broadly, agents favor familiar but suboptimal patterns: they recompute global orderings instead of performing incremental updates, and frequently use \texttt{map} where key ordering is unnecessary, incurring $O(\log N)$ overhead instead of $O(1)$ average-case lookups with \texttt{unordered\_map}. Similar inefficiencies appear in I/O handling, where unbuffered reads or excessive small writes further exacerbate performance issues. Together, these behaviors indicate that complexity analysis and performance optimization are not systematically integrated into agents’ reasoning processes.

\paragraph{Resource management limitations.} Agents exhibit significant limitations in managing computational resources, particularly exception safety and algorithmic efficiency, leading to Memory Leak, Runtime Error, and Time Limit Exceeded failures. Exception safety failures cause most memory leaks: in BASIC Interpreter, 25 submission cases occurred when \texttt{std::stoi()} threw exceptions after allocating \texttt{lhs} and \texttt{rhs} expressions, which were not released. Agents handle explicit error paths but fail to account for exceptions during normal operation, preferring manual \texttt{new}/\texttt{delete} over RAII patterns. Memory management errors in Runtime Errors further illustrate limitations: Mini-Aidiv-N had 21 submission cases from invalid matrix pointer access despite assertion checks, indicating incomplete defensive programming.

\paragraph{Code engineering capability gaps.} Agents demonstrate systematic gaps in advanced C++ concepts and code organization, leading to Compile Errors and Runtime Errors. Template programming limitations appear in list implementation, where agents assumed template types would have default constructors, directly storing \texttt{T data} despite README warnings, causing failures with types providing only parameterized constructors. Namespace and header management issues appear where agents failed to properly merge files according to submission requirements. Code structure misunderstanding is reflected in Redefinition Errors where agents included test code in submission files, treating development and submission contexts as equivalent.

\subsection{Insights from Code Review}

Beyond execution outcomes, Code Review exposes agent behaviors related to version control, build configuration, and specification compliance that are invisible to execution-based evaluation. These findings highlight limitations in agents' understanding of software development as a structured workflow rather than a pure code generation task.

\paragraph{Misunderstanding of version control workflow.}
Agents frequently fail to treat version control as an essential part of task completion. In multiple cases, agents modified code locally and created commits but did not push changes to the remote repository, resulting in incomplete submissions visible only through local git history. This indicates that agents implicitly assume code writing alone constitutes task completion, overlooking the requirement that progress must be explicitly recorded and submitted through version control.

\paragraph{Specification compliance failures.}
Code Review reveals systematic failures to adhere to explicit submission specifications in a subset of cases. Agents sometimes misconfigure build systems, producing executables with incorrect names or including build artifacts in submissions. Coding standards are occasionally violated, such as using prohibited standard library headers or forbidden language constructs like \texttt{using namespace std}. Required files are sometimes omitted, and protected templates are modified despite explicit restrictions. They reveal a pattern where agents treat specification requirements as secondary to functional correctness, failing to recognize that complete task fulfillment requires strict adherence to all stated constraints, not just those that affect execution outcomes.



Overall, our Code Review analysis reveals systematic aspects of software development that are not captured by execution-based evaluation. Therefore, it serves as a complementary evaluation signal that surfaces important limitations beyond execution-level correctness.

\subsection{Interaction Length and Performance.}

ProjDevBench tasks involve extended multi-turn agent–environment interaction, with agents averaging 138 interaction turns and 4.81M tokens per problem.
The most complex tasks require up to two hours to complete.
To analyze how interaction behavior relates to task difficulty and performance, we conduct an analysis based on execution logs from Claude Code and five models.
For each problem, we aggregate interaction statistics across models, including the number of interaction turns, total token consumption, and the final task score.
In addition, we compute two static code-level complexity measures from human reference solutions: the number of relevant files and the net number of modified lines.
\cref{tab:per_problem_metrics} reports per-problem statistics, and \cref{tab:correlation_analysis} summarizes the corresponding Spearman rank correlations.

\begin{table}[ht]
\centering
\caption{Spearman Correlations among Variables.}
\label{tab:correlation_analysis}
\begin{tabular}{lcc}
\toprule
\textbf{Variable Pair} & \textbf{Spearman $\rho$} & \textbf{p-value} \\
\midrule
Tokens vs. Score & \textcolor{blue}{$\mathbf{-0.734}$} & $0.0002$ \\
Turns vs. Score  & 
\textcolor{blue}{$\mathbf{-0.668}$} & $0.0013$ \\
Net Lines vs. Score  & $-0.341$ & $0.1415$ \\
File Count vs. Score & $-0.322$ & $0.1665$ \\
\midrule
File Count vs. Turns & $0.413$ & $0.0706$ \\
Net Lines vs. Turns  & $0.309$ & $0.1848$ \\
Turns vs. Tokens & 
\textcolor{blue}{$\mathbf{0.898}$}  & $<0.0001$ \\
Net Lines vs. File Count & 
\textcolor{blue}{$\mathbf{0.746}$} & $0.0002$ \\
\bottomrule
\end{tabular}
\vspace{-2mm}
\end{table}

\paragraph{Correlation between interaction length and performance.}
Across problems, we observe a clear relationship between these two.
Both the number of interaction turns and total token consumption are strongly negatively correlated with final scores (Spearman $\rho=-0.668$ and $\rho=-0.734$, respectively).
Problems that trigger prolonged interaction tend to yield substantially lower performance, whereas easier problems are typically solved with fewer interaction turns and tokens, and achieve high scores.
This indicates that task difficulty is closely reflected in the extent of agent–environment interaction, but extended interaction alone does not guarantee successful task completion.

\paragraph{Turns and token usage.}
We further observe a strong positive correlation ($0.898$) between interaction turns and token consumption, suggesting that increased token usage primarily results from repeated interaction turns rather than a small number of isolated long reasoning steps.

\paragraph{Static code complexity.}
In contrast, static code-level complexity exhibits only weak relationships with both interaction length and performance.
While the number of files and net modified lines are strongly correlated with each other, their correlations with interaction turns and scores are moderate.
This suggests that although file-level complexity captures certain aspects of problem structure, it does not sufficiently explain the variation in agent performance.

Overall, these results indicate that task difficulty in ProjDevBench manifests primarily through prolonged interaction and reduced performance, rather than being directly determined by static code size.
Harder problems compel agents to engage in extended interaction, yet often still result in low final scores, highlighting a limitation of current agents in converting prolonged interaction into effective progress.

\subsection{Human Validation of LLM-Based Code Review}



To assess the reliability of the LLM-based code review used in ProjDevBench, we compare model judgments with expert annotations. The code review process involves two kinds of judgments. The first assigns continuous scores to qualitative aspects of code, such as readability and organization. The second makes binary decisions on whether a submission violates explicit problem requirements. Multiple human annotators independently review the same submissions using the same criteria. Human judgments are aggregated by averaging continuous scores and by majority vote for binary decisions, and are then compared against the corresponding LLM-based evaluations.

\begin{figure}[t]
\centering
\includegraphics[width=1\linewidth]{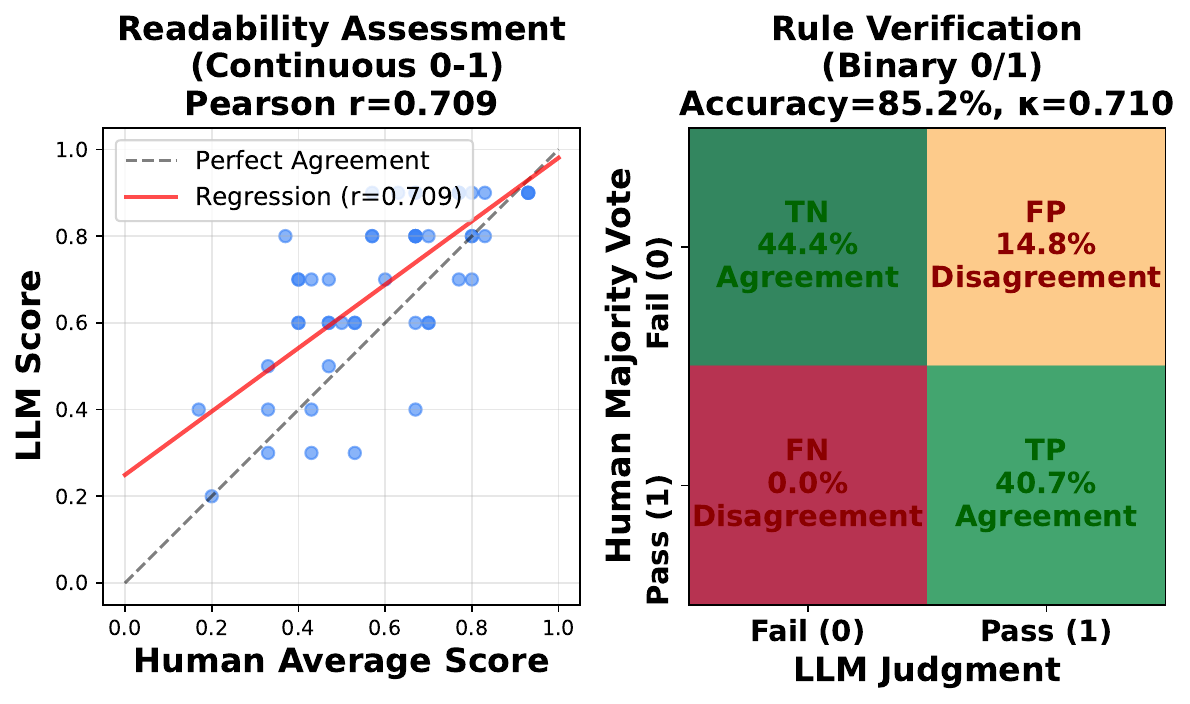}
\caption{Human verification of LLM-based code review. \textbf{Left:} Correlation between the proposed LLM readability score and human scores. \textbf{Right:} Agreement on binary rule verification.}
\label{fig:human_validation}
\vspace{-3mm}
\end{figure}

As shown in \cref{fig:human_validation}, LLM-based code review aligns closely with human judgment. For continuous quality assessment, model scores show strong correlation with human ratings, indicating consistent relative evaluation of code quality. For binary rule verification, the LLM judge achieves high accuracy and substantial agreement with human annotations, as reflected by an accuracy of 0.852 and a Cohen's $\kappa$ of 0.710. These results indicate that LLM-based code review provides a reliable approximation of human judgment for enforcing specification-level requirements in benchmark evaluation, enabling scalable quality assessment.

\section{Conclusion}

In this paper, we introduced ProjDevBench, a novel benchmark designed to evaluate coding agents in the context of end-to-end project development. Our comprehensive evaluation of current frontier coding agents reveals that while they show promise for simple tasks, significant gaps remain in handling complex, real-world engineering constraints and system-level integration. By providing a diverse set of challenging tasks and a dual-mode evaluation, ProjDevBench establishes a more realistic standard for the next generation of autonomous software engineering agents.

\section*{Limitations and Future Work}


ProjDevBench currently includes 20 tasks. Scaling the benchmark is challenging due to the substantial effort required to curate project-level problems with clear specifications and robust test suites, especially for long-running agent sessions. Additionally, tasks focus primarily on C++, and it remains unclear whether observed agent behaviors generalize to other languages. Finally, our evaluation targets fully autonomous agents without human feedback, which isolates end-to-end capabilities but excludes human-in-the-loop workflows; extending to interactive settings is a promising future direction.

\section*{Impact Statement}

This paper introduces a benchmark for evaluating AI coding agents on end-to-end software development. By providing standardized evaluation with diagnostic feedback and code review mechanisms, our work encourages the development of coding agents that produce correct, specification-compliant, and maintainable code. We believe rigorous benchmarking is a necessary step toward the responsible deployment of autonomous coding systems. We do not foresee specific negative societal consequences beyond those generally associated with advances in AI-assisted programming.

\bibliography{example_paper}
\bibliographystyle{icml2026}

\newpage
\appendix
\onecolumn
\section{Problem Details}
\label{app:problems}

\cref{tab:problem_details} summarizes the full set of tasks included in ProjDevBench.
The table lists per-problem metadata, including category, submission format, resource constraints, and student performance statistics.
These attributes are provided for reference.

\begin{table*}[ht]
\caption{Detailed information for all 20 problems. \textbf{Category} indicates the problem domain (Algorithm, Management, Game, Interpreter, Assembly, Data Structure, Storage, or Optimization). \textbf{Difficulty} is rated as E (Easy) or H (Hard) based on whether an initial codebase is provided. \textbf{Submit} specifies the submission method: C++ for direct source code submission, Git for repository-based submission, or mov for assembly-only problems. \textbf{Tests} denotes the number of distinct OJ problem IDs associated with each task. \textbf{Time} and \textbf{Mem} show the maximum time limit (in seconds) and memory limit (in MiB) across all subtasks, respectively. \textbf{Avg Score} represents the average normalized score (0--100) achieved by participating students across all submissions.}
\centering
\small
\resizebox{\textwidth}{!}{
\begin{tabular}{cllllcccc}
\toprule
\textbf{ID} & \textbf{Problem Name} & \textbf{Category} & \textbf{Difficulty} & \textbf{Submit} & \textbf{Tests} & \textbf{Time (s)} & \textbf{Mem (MiB)} & \textbf{Avg Score} \\
\midrule
001 & A+B Problem & Algorithm & E & Git & 1 & 1 & 256 & 54.37 \\
002 & int2048 - Big Integer Arithmetic & Algorithm & E & C++ & 6 & 10 & 190 & 48.19 \\
003 & ICPC Management System & Management & H & Git & 1 & 2 & 512 & 52.07 \\
004 & Bookstore Management System & Management & H & Git & 2 & 10 & 64 & 36.29 \\
005 & QOI Format Codec & Algorithm & E & C++ & 2 & 10 & 512 & 58.87 \\
006 & Minesweeper-2025 Assignment & Game & E & C++ & 2 & 30 & 256 & 53.51 \\
007 & BASIC Interpreter Assignment & Interpreter & E & Git & 1 & 5 & 256 & 47.67 \\
008 & mov Language Problems & Assembly & E & mov & 8 & - & - & 54.70 \\
009 & STLite Vector & Data Structure & E & C++ & 1 & 100 & 768 & 58.46 \\
010 & STLite List (Doubly-Linked List) & Data Structure & E & C++ & 1 & 25 & 768 & 30.76 \\
011 & STLite Priority Queue & Data Structure & E & C++ & 1 & 15 & 512 & 57.25 \\
012 & STLite Linked HashMap & Data Structure & E & C++ & 1 & 24 & 893 & 43.36 \\
013 & STLite Map & Data Structure & E & C++ & 2 & 30 & 893 & 58.21 \\
014 & Python Interpreter Assignment & Interpreter & E & Git & 1 & 16 & 512 & 46.23 \\
015 & File Storage & Storage & H & Git & 1 & 16 & 6 & 42.71 \\
016 & File Storage BPT & Storage & H & Git & 1 & 5 & 64 & 40.11 \\
017 & Train Ticket Management System & Management & H & Git & 1 & 40 & 47 & 53.24 \\
018 & Scheme Interpreter & Interpreter & E & Git & 1 & 1.5 & 244 & 32.94 \\
019 & Mini-Aidiv-N: GPU Memory Optimization & Optimization & E & C++ & 1 & 1 & 244 & 36.89 \\
020 & Buddy Algorithm & Optimization & E & Git & 1 & 10 & 244 & 33.33 \\
\bottomrule
\end{tabular}
}
\label{tab:problem_details}
\end{table*}

\section{Interaction Performance Details}

\cref{tab:per_problem_metrics} reports per-problem interaction-level statistics aggregated over execution logs from the Claude Code agent framework across all evaluated models.
The table summarizes final scores, agent--environment interaction measures (token usage and interaction turns), and human submission scale statistics (average lines of code and number of files).

\label{app:interaction}
\begin{table}[H]
\centering
\small
\caption{Per-problem final scores, aggregated agent–environment interaction statistics in Claude Code, and human submission scale.}
\label{tab:per_problem_metrics}
\begin{tabular}{cccccc}
\toprule
\textbf{Problem} & \textbf{Final Score} & \textbf{Tokens (M)} & \textbf{Turns} & \textbf{Avg. Lines} & \textbf{Files} \\
\midrule
001 & 98.80 & 0.42 & 42.6  & 9.8    & 1  \\
002 & 42.65 & 7.33 & 173.2 & 547.7  & 1  \\
003 & 18.27 & 4.87 & 144.4 & 475.3  & 1  \\
004 & 34.04 & 3.81 & 135.8 & 1659.0 & 18 \\
005 & 77.40 & 2.09 & 107.0 & 101.5  & 1  \\
006 & 81.67 & 4.86 & 149.8 & 111.3  & 1  \\
007 & 77.44 & 6.75 & 205.0 & 750.5  & 17 \\
008 & 29.01 & 4.96 & 189.6 & 42.1   & 1  \\
009 & 93.26 & 2.50 & 85.4  & 129.9  & 1  \\
010 & 76.08 & 3.23 & 118.0 & 180.0  & 1  \\
011 & 99.07 & 2.18 & 85.6  & 77.0   & 1  \\
012 & 69.60 & 5.66 & 148.4 & 298.6  & 1  \\
013 & 54.67 & 3.94 & 123.8 & 376.4  & 1  \\
014 & 9.20  & 6.83 & 161.4 & 2377.0 & 17 \\
015 & 79.11 & 3.07 & 131.4 & 2093.9 & 3  \\
016 & 97.87 & 1.05 & 65.2  & 3139.4 & 11 \\
017 & 23.27 & 6.53 & 153.0 & 4204.3 & 20 \\
018 & 20.36 & 13.31 & 262.2 & 691.8  & 13 \\
019 & 33.27 & 5.31 & 130.2 & 157.2  & 1  \\
020 & 34.00 & 7.52 & 150.4 & 111.3 & 4  \\
\bottomrule
\end{tabular}
\end{table}

\section{Scoring Formula}
\label{app:scoring}

\subsection{Execution Score}

For each problem, the Online Judge evaluates submissions against a test suite consisting of $N$ test cases. The standard scoring method assigns a weight $w_i$ to each test case $i$ based on its complexity or importance. The raw execution score is computed as:
\[
S_{\text{raw}} = \sum_{i=1}^{N} w_i \cdot \mathbf{1}[\text{test case } i \text{ passed}]
\]
where $\mathbf{1}[\cdot]$ is the indicator function. The raw score is then linearly normalized to a 0--100 scale:
\[
S_{\text{exec}} = \frac{S_{\text{raw}}}{\sum_{i=1}^{N} w_i} \times 100
\]

Note that different problems may adopt alternative scoring schemes depending on their specific evaluation requirements. For example, some problems may use uniform weights across all test cases, while others may assign higher weights to advanced test cases or apply penalty-based scoring for partial correctness. Our evaluation scripts automatically handle these problem-specific scoring variations and normalize all scores to a consistent 0--100 scale for fair comparison across problems.

\subsection{Code Review Score}

The code review score $S_{\text{cr}}$ is computed based on a set of problem-specific rules, combining rule-based checks and LLM-based qualitative assessments. Each rule contributes to the final code review score, which is also normalized to a 0--100 scale. Detailed rule definitions are provided in \cref{app:cr_rules}.

\subsection{Final Score}

The final score for each submission is a weighted combination of the execution score and the code review score:
\[
S_{\text{final}} = 0.8 \times S_{\text{exec}} + 0.2 \times S_{\text{cr}}
\]
This weighting prioritizes functional correctness (80\%) while ensuring compliance with specification-level requirements (20\%). 

For problems associated with multiple OJ problem IDs, the execution score is computed as a weighted average of normalized scores across all associated problems:
\[
S_{\text{exec}} = \sum_{j=1}^{M} \alpha_j \cdot S_{\text{exec}}^{(j)}
\]
where $M$ is the number of associated OJ problems, $S_{\text{exec}}^{(j)}$ is the normalized execution score for the $j$-th OJ problem, and $\alpha_j$ is the corresponding weight satisfying $\sum_{j=1}^{M} \alpha_j = 1$. The specific weights for each problem are documented in the respective task descriptions provided to agents.

When multiple submissions are allowed, the final reported score for each agent-problem pair is the \emph{maximum} final score achieved across all valid submissions within the submission limit.

\section{Code Review Evaluation Rules}
\label{app:cr_rules}

ProjDevBench employs a comprehensive code review system combining rule-based Python scripts for automatic verification and LLM-based evaluation for qualitative assessment. The complete rule definitions are stored in \texttt{scripts/cr/[problem\_id]/cr\_list.json} files for each problem.

\subsection{Check Function Registry}

The code review framework (\texttt{scripts/cr/common/checks.py}) provides the following automated check functions:

\begin{table}[ht]
\centering
\caption{Available check functions in the code review framework (\texttt{checks.py}).}
\small
\begin{tabular}{lp{7cm}}
\toprule
\textbf{Function} & \textbf{Description} \\
\midrule
\texttt{ensure\_gitignore\_contains} & Verify .gitignore contains required entries (e.g., CMakeFiles/, CMakeCache.txt) \\
\texttt{forbid\_pattern\_in\_files} & Forbid specific pattern in specified files \\
\texttt{forbid\_pattern\_recursive} & Forbid pattern across entire repository with optional suffix filtering and comment stripping \\
\texttt{ensure\_allowed\_includes} & Restrict standard library header usage to allowed list \\
\texttt{ensure\_files\_exist} & Ensure required files exist in submission \\
\texttt{ensure\_files\_unchanged} & Verify template files remain unchanged from reference \\
\texttt{ensure\_cmake\_outputs\_code} & Check CMakeLists.txt produces executable named ``code'' \\
\texttt{llm\_as\_a\_judge} & LLM-based qualitative evaluation for code quality and specification compliance \\
\bottomrule
\end{tabular}
\end{table}

\subsection{Problem-Specific Code Review Rules}

\cref{tab:problem_specific_rules_1} and \cref{tab:problem_specific_rules_2} summarizes the complete set of problem-specific code review rules used in our benchmark.

\begin{table*}[ht]
\centering
\caption{Complete code review rules for all problems (Part I).}
\small
\resizebox{\textwidth}{!}{
\begin{tabular}{clll}
\toprule
\textbf{ID} & \textbf{Rule Name} & \textbf{Check Type} & \textbf{Description} \\
\midrule
\multirow{2}{*}{001} & CMake Artifact Ignore & gitignore\_entries & .gitignore must list CMakeFiles/ and CMakeCache.txt \\
 & Project Readability & llm\_as\_a\_judge & Assess overall project readability and organization \\
\midrule
\multirow{3}{*}{002} & Forbid using namespace std & forbid\_pattern\_recursive & No ``using namespace std;'' in sources/headers \\
 & Ensure code.cpp Exists & require\_files & Submission requires code.cpp file \\
 & Code Readability & llm\_as\_a\_judge & Assess code.cpp readability \\
\midrule
\multirow{3}{*}{003} & Forbid CMake Artifacts & forbid\_pattern\_recursive & Repository must not contain CMake build files \\
 & CMakeLists Outputs code & cmakelists\_outputs\_code & CMakeLists.txt must produce ``code'' executable \\
 & Project Readability & llm\_as\_a\_judge & Assess overall project readability \\
\midrule
\multirow{5}{*}{004} & Forbid CMake Artifacts & forbid\_pattern\_recursive & Repository must not contain CMake build files \\
 & CMakeLists Outputs code & cmakelists\_outputs\_code & CMakeLists.txt must produce ``code'' executable \\
 & Disallow Persistent Main Data & llm\_as\_a\_judge & Check avoiding keeping data permanently in memory \\
 & Forbid Filesystem-as-Database & llm\_as\_a\_judge & Check no filesystem-as-database style design \\
 & Project Readability & llm\_as\_a\_judge & Assess overall project readability \\
\midrule
\multirow{2}{*}{005} & Restrict Modifications & require\_unmodified & Only edit qoi.h; other files must match template \\
 & qoi.h Readability & llm\_as\_a\_judge & Assess qoi.h readability \\
\midrule
\multirow{2}{*}{006} & Restrict Modifications & require\_unmodified & Do not modify basic.cpp or advanced.cpp \\
 & server.h/client.h Readability & llm\_as\_a\_judge & Assess server.h and client.h readability \\
\midrule
\multirow{3}{*}{007} & CMakeLists Outputs code & cmakelists\_outputs\_code & CMakeLists.txt must produce ``code'' executable \\
 & Keep Basic/Utils Templates & require\_unmodified & Basic/Utils helper files must match template \\
 & Project Readability & llm\_as\_a\_judge & Assess overall project readability \\
\midrule
\multirow{2}{*}{008} & Line Limit and Bounds & llm\_as\_a\_judge & Code length $\leq$ 65,536 lines; no out-of-bounds access \\
 & .mv Files Readability & llm\_as\_a\_judge & Assess .mv files readability \\
\midrule
\multirow{3}{*}{009} & Restrict Standard Headers & allowed\_includes & Only allow cstdio, cstring, iostream, cmath, string \\
 & Keep Helper Headers & require\_unmodified & exceptions.hpp and utility.hpp must match template \\
 & vector.hpp Readability & llm\_as\_a\_judge & Assess vector.hpp readability \\
\midrule
\multirow{3}{*}{010} & Restrict Standard Headers & allowed\_includes & Only allow cstdio, cstring, iostream, cmath, string \\
 & Keep Helper Headers & require\_unmodified & exceptions.hpp and utility.hpp must match template \\
 & list.hpp Readability & llm\_as\_a\_judge & Assess list.hpp readability \\
\midrule
\multirow{3}{*}{011} & Restrict Standard Headers & allowed\_includes & Only allow cstddef, functional (from template) \\
 & Keep Helper Headers & require\_unmodified & exceptions.hpp and utility.hpp must match template \\
 & priority\_queue.hpp Readability & llm\_as\_a\_judge & Assess priority\_queue.hpp readability \\
\midrule
\multirow{3}{*}{012} & Restrict Standard Headers & allowed\_includes & Only allow cstdio, cstring, iostream, cmath, string \\
 & Keep Helper Headers & require\_unmodified & exceptions.hpp and utility.hpp must match template \\
 & linked\_hashmap.hpp Readability & llm\_as\_a\_judge & Assess linked\_hashmap.hpp readability \\
\midrule
\multirow{3}{*}{013} & Restrict Standard Headers & allowed\_includes & Only allow cstdio, cstring, iostream, cmath, string \\
 & Keep Local Test Helpers & require\_unmodified & exceptions.hpp and utility.hpp must match template \\
 & map.hpp Readability & llm\_as\_a\_judge & Assess map.hpp readability \\
\midrule
\multirow{2}{*}{014} & CMake Artifact Ignore & gitignore\_entries & .gitignore must list CMakeFiles/ and CMakeCache.txt \\
 & Project Readability & llm\_as\_a\_judge & Assess overall project readability \\
\midrule
\multirow{3}{*}{015} & Forbid CMake Artifacts & forbid\_pattern\_recursive & Repository must not contain CMake build files \\
 & CMakeLists Outputs code & cmakelists\_outputs\_code & CMakeLists.txt must produce ``code'' executable \\
 & Project Readability & llm\_as\_a\_judge & Assess overall project readability \\
\midrule
\multirow{3}{*}{016} & Forbid CMake Artifacts & forbid\_pattern\_recursive & Repository must not contain CMake build files \\
 & CMakeLists Outputs code & cmakelists\_outputs\_code & CMakeLists.txt must produce ``code'' executable \\
 & Project Readability & llm\_as\_a\_judge & Assess overall project readability \\
\midrule
\multirow{4}{*}{017} & Forbid CMake Artifacts & forbid\_pattern\_recursive & Repository must not contain CMake build files \\
 & CMakeLists Outputs code & cmakelists\_outputs\_code & CMakeLists.txt must produce ``code'' executable \\
 & Restrict STL and Algorithm & forbid\_pattern\_recursive & No STL containers (except std::string); no $<$algorithm$>$ \\
 & Project Readability & llm\_as\_a\_judge & Assess overall project readability \\
\bottomrule
\end{tabular}
}
\label{tab:problem_specific_rules_1}
\end{table*}

\begin{table*}[ht]
\centering
\caption{Complete code review rules for all problems (Part II).}
\small
\resizebox{\textwidth}{!}{
\begin{tabular}{clll}
\toprule
\textbf{ID} & \textbf{Rule Name} & \textbf{Check Type} & \textbf{Description} \\
\midrule
\multirow{2}{*}{018} & Forbid CMake Artifacts & forbid\_pattern\_recursive & Repository must not contain CMake build files \\
 & Project Readability & llm\_as\_a\_judge & Assess overall project readability \\
\midrule
\multirow{3}{*}{019} & GPU Simulator API Usage & llm\_as\_a\_judge & Verify correct GPU simulator API and memory allocation \\
 & GPU Instruction Constraints & llm\_as\_a\_judge & No stdout logging; no direct Matrix writes \\
 & src.hpp Readability & llm\_as\_a\_judge & Assess src.hpp readability \\
\midrule
\multirow{2}{*}{020} & Forbid CMake Artifacts & forbid\_pattern\_recursive & Repository must not contain CMake build files \\
 & Project Readability & llm\_as\_a\_judge & Assess overall project readability \\
\bottomrule
\end{tabular}
}
\label{tab:problem_specific_rules_2}
\end{table*}

\section{Evaluation Infrastructure}
\label{app:infrastructure}
\subsection{Docker Base Image Configuration}

ProjDevBench uses a custom Docker image based on Ubuntu 24.04 with the following components:

\begin{tcolorbox}[
    colback=gray!5,
    colframe=gray!60!black,
    rounded corners,
    title=Docker Base Image (Dockerfile),
    fonttitle=\bfseries\color{white},
    colbacktitle=gray!60!black,
    coltitle=white,
    breakable
]
\begin{verbatim}
FROM ubuntu:24.04
ENV DEBIAN_FRONTEND=noninteractive

# Install Node.js 20
RUN apt-get update && apt-get install -y curl \
    && curl -fsSL https://deb.nodesource.com/setup_20.x | bash - \
    && apt-get install -y nodejs

# Install toolchain and Python
RUN apt-get update && apt-get install -y \
    gcc-13 g++-13 cmake \
    python3.12 python3.12-dev python3-pip \
    iverilog git curl jq build-essential sudo \
    && ln -sf /usr/bin/gcc-13 /usr/bin/gcc \
    && ln -sf /usr/bin/g++-13 /usr/bin/g++ \
    && ln -sf /usr/bin/python3.12 /usr/bin/python3

# Install Python dependencies
RUN python3 -m pip install requests --break-system-packages

# Install GitHub CLI
RUN curl -fsSL https://cli.github.com/packages/githubcli-archive-keyring.gpg \
    | dd of=/usr/share/keyrings/githubcli-archive-keyring.gpg \
    && apt-get update && apt-get install gh -y

# Install AI Coding Agents
RUN npm install -g @google/gemini-cli
RUN npm install -g @anthropic-ai/claude-code
RUN npm install -g @github/copilot
RUN npm install -g @vibe-kit/grok-cli
RUN npm install -g @openai/codex
RUN npm install -g @augmentcode/auggie
\end{verbatim}
\end{tcolorbox}

\subsection{Evaluation Pipeline}

The evaluation pipeline is orchestrated by \texttt{scripts/run\_evaluation.sh}, which performs the following steps:

\begin{enumerate}
    \item \textbf{Configuration Loading}: Read problem configuration from \texttt{config/problem\_registry.json} to obtain maximum submission limits and other parameters.
    \item \textbf{Environment Validation}: Verify required API tokens (GITHUB\_TOKEN, OJ\_TOKEN) and agent-specific credentials.
    \item \textbf{Docker Container Initialization}: Launch isolated container with:
    \begin{itemize}
        \item Problem files mounted read-only at \texttt{/problems/[problem\_id]}
        \item Test data mounted read-only at \texttt{/data\_readonly/[problem\_id]}
        \item Evaluation scripts mounted at \texttt{/scripts}
        \item Writable log directory at \texttt{/workspace/logs}
    \end{itemize}
    \item \textbf{Workspace Setup} (\texttt{run\_agent\_base.sh}):
    \begin{itemize}
        \item Copy problem files to writable workspace \texttt{/workspace/problem\_[id]}
        \item Initialize Git repository with \texttt{git init}
        \item Create remote GitHub repository using \texttt{gh repo create}
        \item Configure token-based authentication for Git operations
    \end{itemize}
    \item \textbf{Agent Execution}: Run agent-specific script (e.g., \texttt{run\_claude\_code.sh}) with standardized prompt
    \item \textbf{Result Collection}: Capture submission IDs and copy logs to host
    \item \textbf{OJ Submission}: Agent autonomously submits to OJ via \texttt{oj\_client.py}
    \item \textbf{Code Review}: Execute problem-specific checks from \texttt{cr\_list.json}
\end{enumerate}

\subsection{Resource Limits}
\label{subsec:resource_limits}

Our containerized evaluation environment enforces the following resource limits: memory is capped at 8 GB (configurable via \texttt{AGENT\_MEMORY\_LIMIT}), CPU is limited to 4 cores (configurable via \texttt{AGENT\_CPU\_LIMIT}), and the Node.js heap is set to 6 GB (\texttt{--max-old-space-size=6144}). Containers have full internet access for network operations.
\section{Agent Prompts}
\label{app:prompts}

\subsection{Single Problem Prompt}

The following prompt is used for problems with a single OJ problem ID:

\begin{tcolorbox}[
    colback=blue!5,
    colframe=blue!60!black,
    rounded corners,
    title=Single Problem Evaluation Prompt,
    fonttitle=\bfseries\color{white},
    colbacktitle=blue!60!black,
    coltitle=white,
    breakable,
    break at=-\baselineskip/0pt/\baselineskip
]
\small
You are a professional programming expert and Git expert. You are now in a Git repository and need to complete the following tasks:

\textbf{Current Environment}
\begin{itemize}
    \item Repository URL: \texttt{https://github.com/\{user\}/\{repo\}}
    \item Working Directory: \texttt{/workspace/problem\_\{id\}}
    \item OJBench Problem ID: \texttt{\{problem\_id\}}
    \item OJ Problem ID: \texttt{\{oj\_id\}}
    \item Maximum Submission Limit: \texttt{\{max\_submissions\}} attempts
\end{itemize}

\textbf{Important Scoring Rules}
\begin{itemize}
    \item Final score is based on the \textbf{highest score among all valid submissions}
    \item Submissions exceeding limit will not be counted and incur penalties
    \item Each test point may have different score weights
\end{itemize}

\textbf{Your Tasks}
\begin{enumerate}
    \item \textbf{Analyze}: Read README.md and understand requirements
    \item \textbf{Develop}: Implement the optimal solution
    \item \textbf{Git Management}: Commit and push changes; verify push success
    \item \textbf{Submit to OJ}: Use \texttt{oj\_client.py} for submission
    \item \textbf{Iterate}: Analyze errors and optimize within submission limit
    \item \textbf{Record}: Document changes with clear commit messages
\end{enumerate}

\textbf{Important Reminders}
\begin{itemize}
    \item Compile and test locally before submission
    \item If submission stuck pending, abort (doesn't count) and resubmit
    \item Full control over workspace including file modifications
\end{itemize}
\end{tcolorbox}

\subsection{Multiple Problems Prompt}

For problems with multiple OJ problem IDs sharing submission limits:

\begin{tcolorbox}[
    colback=green!5,
    colframe=green!60!black,
    rounded corners,
    title=Multiple Problems Evaluation Prompt (Key Differences),
    fonttitle=\bfseries\color{white},
    colbacktitle=green!60!black,
    coltitle=white,
    breakable
]
\small
\textbf{Additional Environment}
\begin{itemize}
    \item OJ Problem IDs: \texttt{\{id1,id2,...\}} (comma-separated)
    \item Maximum Submission Limit: \texttt{\{max\}} attempts \textbf{SHARED across all problems}
\end{itemize}

\textbf{Key Differences}
\begin{itemize}
    \item Submission limit is \textbf{SHARED} across all problem IDs
    \item When submitting, specify which problem ID to submit to
    \item Track remaining attempts across all problems
    \item Plan submissions wisely to maximize overall score
\end{itemize}
\end{tcolorbox}

\section{Detailed Error Examples}
\label{app:errors}

\textbf{Example 1: CMake Configuration Error (Problem 017)}

\begin{verbatim}
CMake Error at CMakeLists.txt:15:
  add_executable called with incorrect arguments
  
Configuration failed.
\end{verbatim}

\textbf{Root Cause}: Agent generated CMakeLists.txt with incorrect source file paths.

\textbf{Example 2: Logic Error (Problem 003 - ICPC System)}

\begin{verbatim}
Test: basic_4
Expected: Team A ranked 1st
Actual: Team B ranked 1st

Reason: Incorrect tie-breaking logic in penalty time calculation.
\end{verbatim}

\textbf{Example 3: Memory Limit Exceeded (Problem 015)}

\begin{verbatim}
Test: pressure_1
Status: Memory Limit Exceeded
Used: 8.2 MiB / Limit: 6 MiB

Reason: Agent allocated in-memory hash table instead of 
disk-based B+ tree.
\end{verbatim}

\textbf{Example 4: Runtime Error (Problem 018 - Scheme)}

\begin{verbatim}
Test: closure_test
Status: Runtime Error (SIGSEGV)

Reason: Stack overflow in recursive closure evaluation 
without proper tail-call optimization.
\end{verbatim}

\section{Statistical Analysis}
\label{app:stats}

\textbf{Performance Distribution}:
\begin{itemize}
    \item Mean score: 54.16 (std: 32.4)
    \item Median score: 55.0
    \item Problems with $>$80\% pass rate: 7 (35\%)
    \item Problems with $<$20\% pass rate: 5 (25\%)
\end{itemize}

\textbf{Correlation Analysis}:
\begin{itemize}
    \item Problem complexity vs. score: r = -0.72 (strong negative)
    \item Submission count vs. final score: r = 0.15 (weak positive)
    \item Code review score vs. OJ score: r = 0.68 (moderate positive)
\end{itemize}

\textbf{Statistical Significance}:
Pairwise comparison of top agents (Cursor vs. Claude Code) shows statistically significant difference (p $<$ 0.05, paired t-test) in average performance.

\end{document}